\documentclass[11pt,a4paper]{article}
\usepackage[hyperref]{acl2018}
\usepackage{times}
\usepackage{latexsym}
\usepackage{capt-of}
\usepackage{url}
\usepackage{graphicx}

\usepackage{url}

\aclfinalcopy 

\newcommand{\me}{\textit{M}}
\newcommand{\sd}{\textit{SD}}

\title{``Can you say more about the location?'' \\The Development of a Pedagogical Reference Resolution Agent}

\author{Maike Paetzel \\
  Social Robotics Lab \\
  Uppsala University \\
  Sweden \\
  {\tt maike.paetzel@it.uu.se} \\\And
  Ramesh Manuvinakurike \\
  Institute for Creative Technologies \\
  University of Southern California \\
  United States of America \\
  {\tt manuvina@usc.edu} \\}

\date{}

\begin{document}
\maketitle
\begin{abstract}
In an increasingly globalized world, geographic literacy is crucial. In this paper, we present a collaborative two-player game to improve people's ability to locate countries on the world map. We discuss two implementations of the game: First, we created a web-based version which can be played with the remote-controlled agent Nellie. With the knowledge we gained from a large online data collection, we re-implemented the game so it can be played face-to-face with the Furhat robot Neil. Our analysis shows that participants found the game not just engaging to play, they also believe they gained lasting knowledge about the world map. 

\end{abstract}

\section{Introduction}
\label{sec:introduction}

\begin{table*}[ht]
\begin{minipage}{0.3\linewidth}
\vspace{0pt}
\small
\begin{tabular}{  l  p{4.4cm}  }
    \hline
    Role & utterance \\ \hline 
    Dir &  The first country is in middle of Africa. It's South Sudan. Do you happen to know where that is? \\ 
    Mat & I don't know where that is \\ 
    Dir & That's surprising \$laughter\$ \ldots Do you know where Egypt is?\\ 
    Mat & No \\ 
    Dir & Look at the um Africa \ldots top three biggest countries uh on top of Africa \ldots You see those? \\ 
    Mat & Yes \\ 
    Dir & The one to the furthest right \ldots is Egypt\\ 
    Mat & Ok \\ 
    Dir & Go two down \ldots that is South Sudan \\ 
    Mat & What does it look like? \\ 
    Dir & \ldots Kinda looks like seahorse laying on its back \ldots \\
    Mat & Got it \\ \hline
   \end{tabular}
\end{minipage}
\begin{minipage}{0.7\linewidth}
\centering
\includegraphics[width=3.21in, height=3.1in]{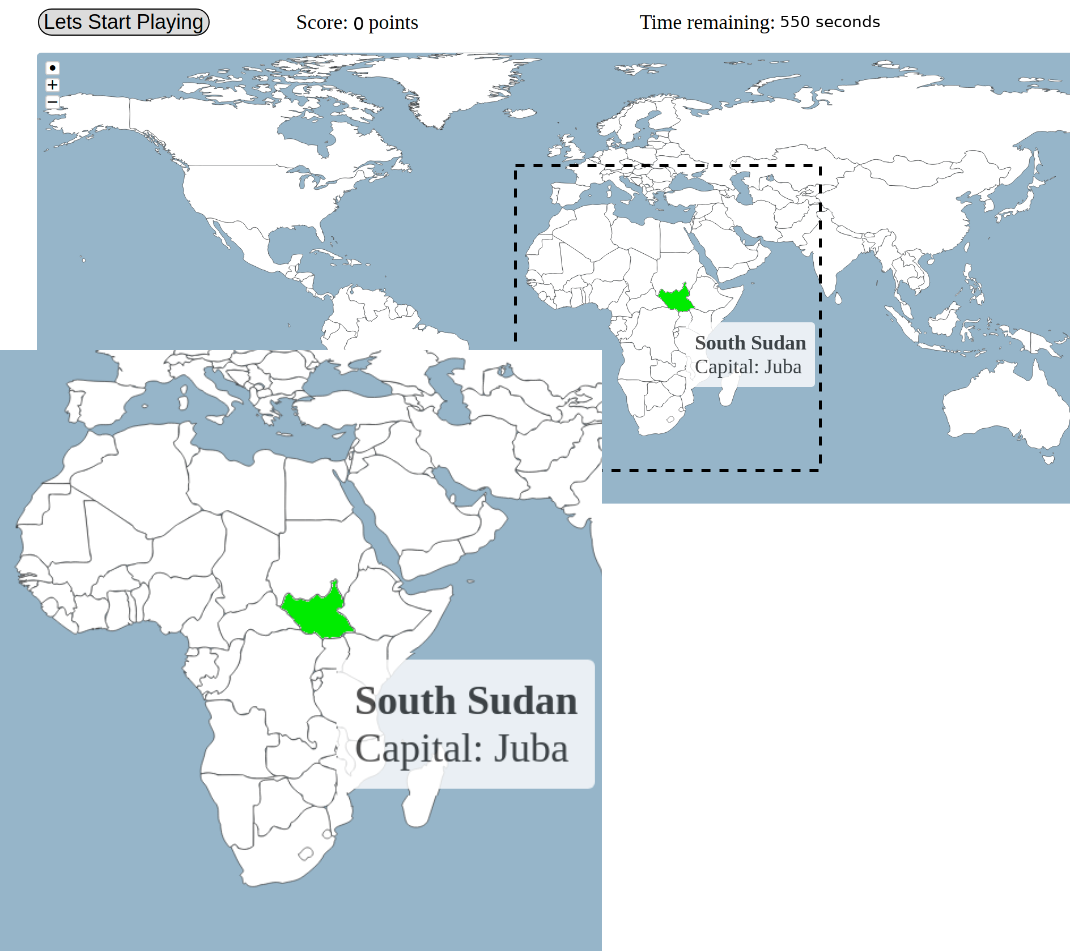}
\label{fig:map-dir}
\end{minipage}

\captionof{figure}{The map as presented to the Director. The target country is highlighted in green and information about the country are presented when hovering over it (highlighted in gray). The table to the left shows a sample conversation between the human Director (Dir) and the agent Matcher (Mat).}
\end{table*}

In an increasingly connected and globalized world, knowledge about other countries is not only crucial in many jobs but also to make informed decisions about which political agenda to support or to understand and build an opinion about a geo-political news articles. 
However, in 2006, the National Geographic Society found that ``young people in the United States - the most recent graduates of our educational system - are unprepared for an increasingly global future''  \cite{natgeo:2006}. Even though 78\% of survey respondents found it at least somewhat important to know where countries in the news are located, six in ten could not find Iraq on a map despite almost constant news coverage. In an even more recent survey \cite{natgeo:2016}, it was found that university students could locate only 46.5\% of the countries that the US is actively engaged politically on the map. 
The National Geographic Society suggests that the level of education plays a major role in geographic literacy. In other words, ``the more education respondents have, the more likely they are to answer questions on geographic literacy correctly''  \cite{natgeo:2006}. However, many Americans do not progress their education beyond high school, which leads to the question: \textit{How can we educate those lacking geographic literacy skills independent of their educational level?} 

Spoken dialogue system (SDS) technologies have made rapid progress in the past decade: Automatic speech recognizers with low word error rates, machine learning approaches that yield high accuracy speech act labels for language understanding and speech synthesizers that create human-like voices have made it possible to develop numerous new SDS applications. One such application are \textit{Intelligent Tutoring Systems}. 
Over the past few decades, Intelligent Tutoring Systems  have been shown to successfully achieve appreciable learning gains (cf~\newcite{ lesgold1988sherlock, freedman1999atlas, koedinger1997intelligent, mitrovic1999evaluation, gertner2000andes,graesser2001intelligent, litman2004itspoke, graesser2004autotutor, mcnamara2004istart, craig2013impact,koedinger2013using, pane2014effectiveness, graesser2016conversations,trinh2017robocop}). 
Those systems usually rely heavily on experts hand-authoring the pedagogical content. 
While this ensures high quality, it also makes such systems expensive and time-consuming to build. To overcome this issue, crowd-sourcing content creation and quality control have shown potential in recent times  \cite{mitros2014creating, baker2016stupid}. While pedagogical tutoring systems generally give the user much freedom in how to use the system, the increased pedagogic value often comes at the cost of decrease in the fun and engagement with the system. 

Another class of pedagogical systems are \textit{Serious Games} \cite{michael2005serious, johnson2005serious}. They are typically developed along the narrative of a storyboard which leads the learner to solve problems, which results in learning gains. 
These games usually don't struggle to engage the learner; however, the hand-authored storyboard limits their options to influence the story significantly. 
This may decrease the learning gain since the users don't have to find potential problem solutions themselves. 
Another drawback of Serious Games is that, since the games are story-based, repeated interactions with these games are often not intended. 

The main contribution of this work is the development of a \textit{Serious Dialogue Game} which aims to (i) maximize the learner's engagement and learning even in repeated interactions with open-ended spoken interactions with the system, while (ii) abstaining entirely from involving experts in the development of the learning content. We picked the skill of locating a country on the world map as an example for our collaborative two-player game. In this work, we focus on presenting the domain and discussing how it serves the purpose of creating high engagement while maintaining a subjectively high learning gain. 
The ultimate goal of this project is to develop an autonomous team partner for the game. 
As a step towards this goal, we developed a Wizard-of-Oz framework that allows studying people's game interactions with an agent.

We developed this game to operate web-based and with an embodied agent as a team partner. Since 90\% of adults in the U.S use the internet \cite{anderson201913}, a web-based version is a good option to reach the general public. Web-based games scale easily, don't have demanding hardware requirements, and potentially allow multiple people playing at the same time. Players can also access them in a safe space, which can decrease the fear of judgment or stigma. 
In addition to the web-based version, we aim to explore how the game could be used in a classroom setting. 
It has been shown that embodied interactions, especially in tutoring systems, can increase the learning gain and the level of engagement \cite{mcnamara2010intelligent}. 
Such systems could eventually be used to enrich the geography lessons in schools. 
Thus, we also implemented a version of the game that can be played together with a Furhat robot.

\section{The RDG-Map Game}
\label{sec:rdgmap}

  \begin{figure*}[t]
    \begin{center} 
    \includegraphics[width=0.48\linewidth]{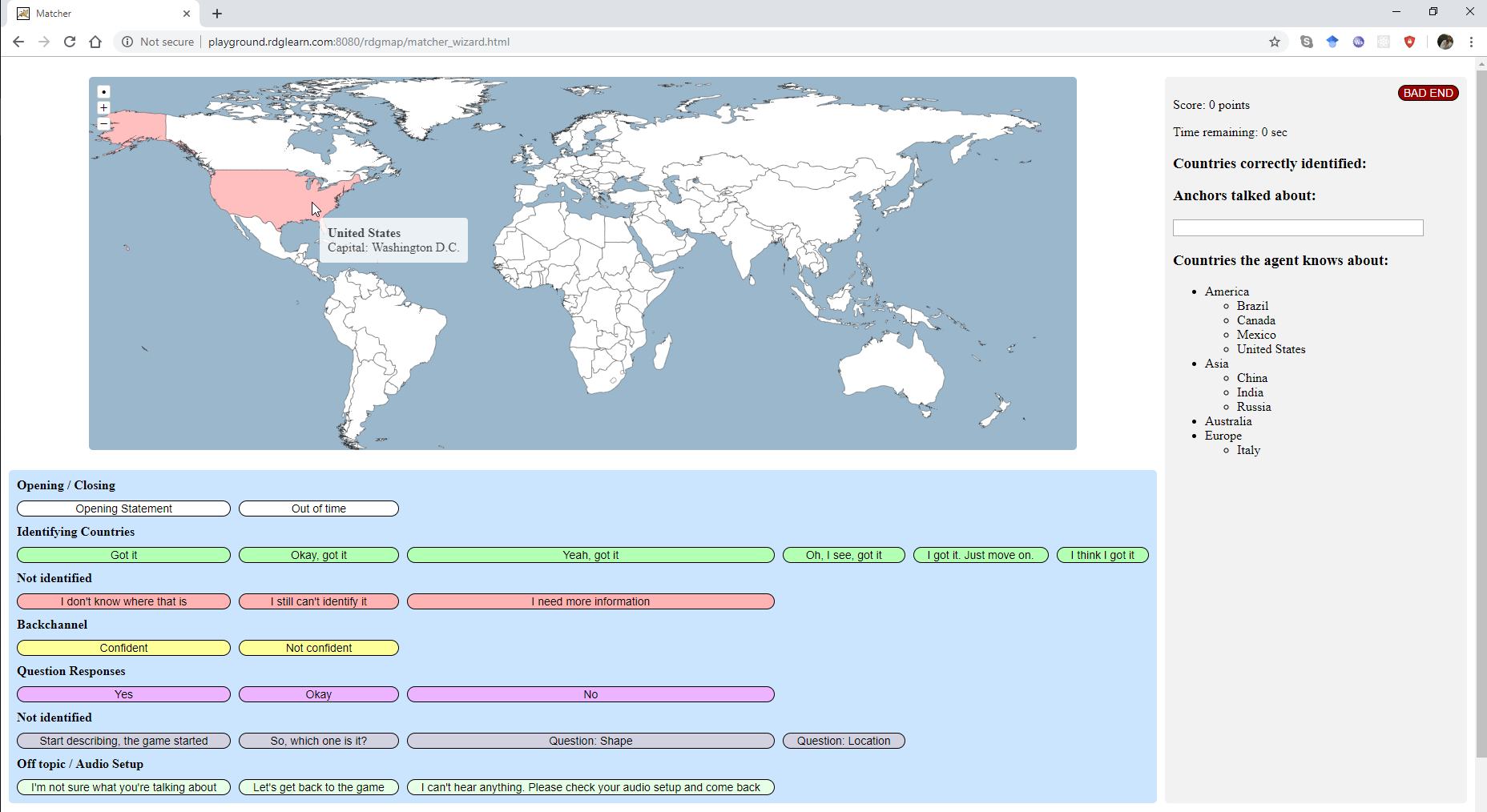}
    \hspace{0.1cm}
    \includegraphics[width=0.48\linewidth]{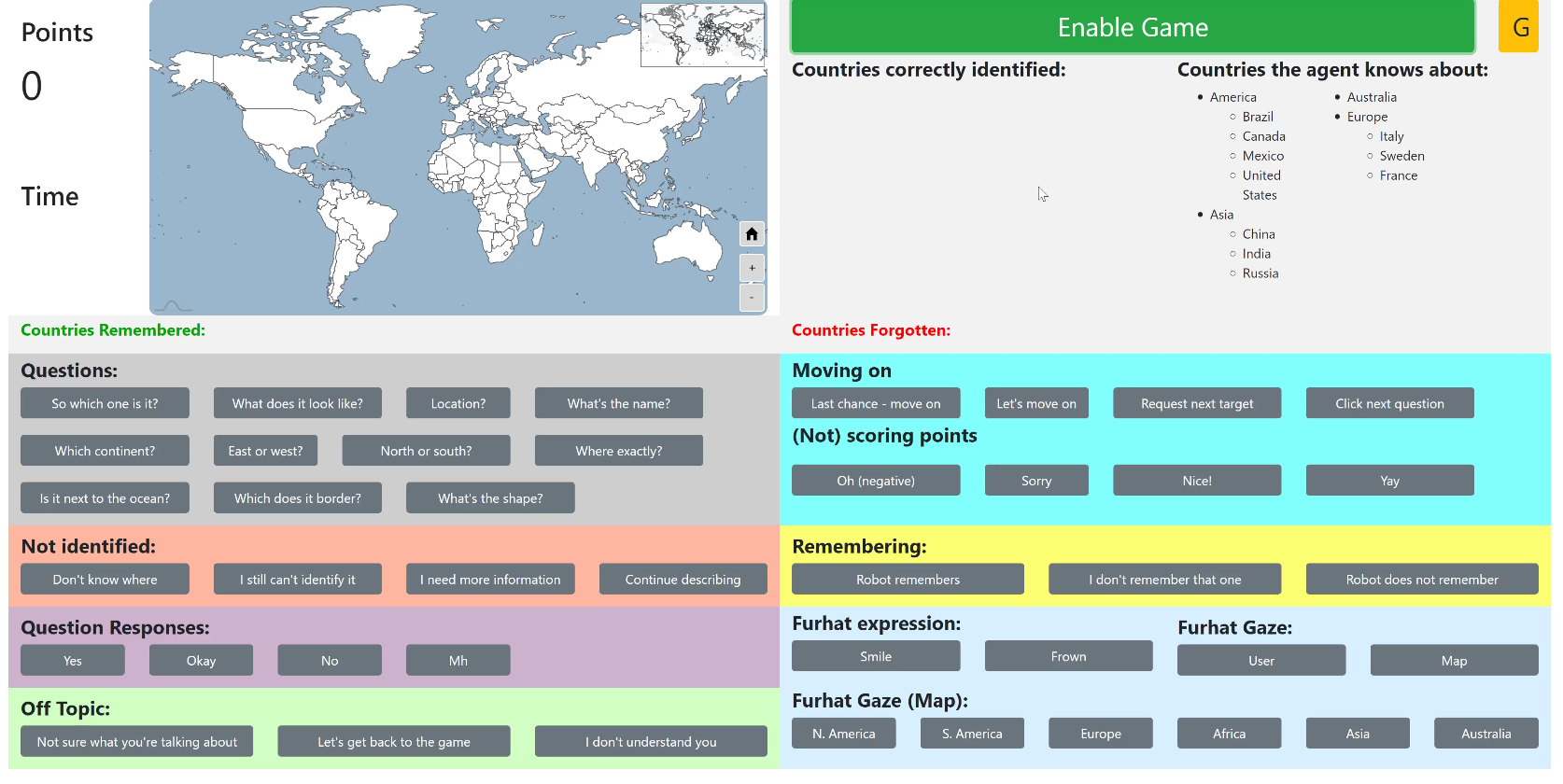}
    
    \caption{The wizard interface for the web-based (left) and the embodied data collection (right).}
    \label{fig:wizardinterface}
    \end{center}
  \end{figure*}

The game we present in this work is developed to promote learning of the size, location, and relation of countries in the world and is called RDG-Map (Rapid Dialogue Game - Map). Learning shall be achieved by playing an interactive, collaborative, and time-constrained two-person game, in which one player is provided with more information and thereby implicitly acts as a tutor in the learning task. However, the aim is that players in both roles can gain knowledge from playing the game. 

The game developed in this work also shares similarity with numerous reference resolution games developed in recent times \cite{paetzel2014multimodal,manuvinakurike2016real,zarriess2016pentoref, de2017guesswhat}. One of the players is assigned the role of a \textit{Director} and the other the role of a \textit{Matcher}. Both players see a map of the world on their respective screen. One of the countries is randomly selected as a target country (TC) and highlighted on the Director's screen in green, as shown in Figure \ref{fig:map-dir}. In addition, the Director may hover over a country to see the name of the country. 
The goal of the Director is to describe the TC so that the Matcher is able to select the same country on their map. The Director and Matcher are able to talk back-and-forth freely to identify the TC. This includes the usage of the countries name. However, the map of the Matcher is not labeled, so the name of the country alone is likely not sufficient to identify the TC. The Matcher can change their selection as often as desired, but the information which country the Matcher has currently selected is  not shared with the Director.

When the Matcher believes (s)he has clicked on the correct country, (s)he communicates this to the Director who presses a button to request the next TC. The team scores a point for each correct guess, with the goal to score as high as possible in the given 600 seconds game time. 

\section{Nellie - A Web-Based Agent}
\label{sec:nellie}

We first implemented a browser-based version of the RDG-Map game that can be either played by a human-human or a human-agent team. 
Communication between two human players is realized using the HTML5 Simplewebrtc\footnote{https://www.simplewebrtc.com/} tool. 

\subsection{Agent Implementation}
\label{subsec:nellie:agentimplementation}
  
We created a Wizard-of-Oz setup in which a human director was paired with what (s)he believed to be an autonomous agent Matcher called Nellie. The Wizard interface was inspired by some initial human-human data collection and the utterances of the human Matcher in the games we analyzed. The interface is divided into three sections (cf. Figure~\ref{fig:wizardinterface} left): On top, the wizard sees the world map, and by clicking on a country, the wizard controls the agent's country selection. On the right, the wizard sees the country names the agent knows by name (the human director is not aware of the countries known by the Matcher). Initially, the agent can locate nine countries on the world map (United States, Canada, Mexico, Brazil, India, China, Russia, Australia, and Italy). These represent all the countries more than 50\% of Americans can find on the world map \cite{natgeo:2006}. 
Whenever a country was correctly identified, the name is automatically added to the panel. In addition, the wizard can add countries or other anchoring points (e.g.: Egypt in Figure~\ref{fig:map-dir}) manually. 
On the lower part of the Wizard screen, 23 thematically themed buttons control the agent's utterances. 
These utterances include opening and closing statements, confirmations, backchannels, yes/no responses, and other game-related statements and questions. \newcite{Cereproc}'s voice Kate is used for agent's voice. The wizard controlling the agent followed the same strict guidelines for every game interaction to ensure the behavior would be as close as possible to the desired behavior of the future autonomous agent.

\subsection{Data Collection}
\label{subsec:nellie:datacollection}

\begin{figure*}[t]
    \begin{center} 
    \includegraphics[width=\linewidth]{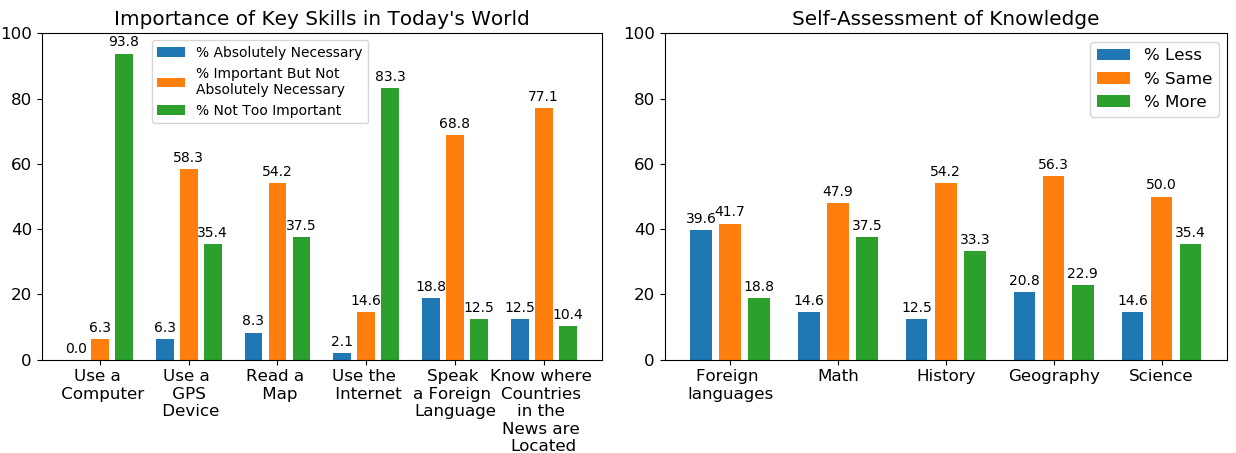}
    
    \caption{The perceived importance of key skills in today's world and the self-assessed knowledge of different subjects by the 48 participants in the web-based data collection.}
    \label{fig:demographics}
    \end{center}
\end{figure*}

The architecture of the system is similar to the one we previously developed and described in \newcite{manuvinakurike2015pair, manuvinakurike2015reducing}. In this work, we extended the framework to support human-wizard interactions. 
Prior to starting the game, instructions were provided as text and a video demonstration. 
Participants were also asked for consent and to fill out a demographic questionnaire. As the framework supports only one Wizard, participants had to wait in a FIFO (First in first out) queue until the wizard was ready to play. 
After participants were paired with the wizard and finished the game, they were directed to a post-questionnaire in which they were asked to reflect on their satisfaction with the agent's performance and the overall gameplay. 

50 native English speakers located in the US were recruited on Amazon Mechanical Turk (MTurk) to play the game . Participants received monetary compensation for their participation. We excluded two game interactions from the analysis, one due to audio problems and the other because the participant did not complete the post-game questionnaire. Out of the 48 remaining participants (Age: \me = 35.7, \sd = 9.7), 19 were female, all had at least completed high school, and 13 were currently or prior enrolled in Computer Science or a related University course.  

In  the demographics questionnaire, we included relevant questions from the \newcite{natgeo:2006} survey to understand the geographic literacy of our participants (cf~Figure \ref{fig:demographics}). 
While the majority of participants agreed that being able to use a Computer (93.8\%) and the Internet (83.3\%) are both absolutely necessary skills, only 12.5\% found it necessary to be able to speak a foreign language and 10.4\% to know where countries in the news are located. 
Participants rated their own skills worst when it comes to foreign languages (39.6\% believe they know less than the average person), followed by geography skills (20.8\% believe they know less than the average person). This is in line with the 2016 findings by the National Geographic Society \cite{natgeo:2006}.

\section{Neil - An Embodied Agent}
\label{sec:neil}

With the experience we gained from the web-based data collection, we revised the game so it could be played in cooperation with the robot head Furhat \cite{c5}. We ultimately aim to create a classroom setting, in which part of the geography curriculum could eventually involve playing the RDG-Map game. As a first step towards this goal, we conducted a Wizard-of-Oz experiment with University students.

\subsection{Game Implementation}
\label{subsec:neil:gameimplementation}

In comparison to the web-based version of the game, the game logic needed changes in order to incorporate the physical embodiment of the agent. While the web-based game was played on two separate screens, we aimed to add an element of shared attention to make more use of the physical presence of the agent. The actual implementation was inspired by a comment from a participant in the web-based data collection who wrote: ``I think it might be better if I was able to see if partner had picked the right countries.'' We often noticed frustration with the agent when they didn't score a point and the human Director could not comprehend the reason for the miss. 
With seeing the selection of the agent, we believed the Director would be faster in distinguishing good and helpful from poor clues. The Director would also be able to correct own mistakes, eg. when confusing East and West or Africa and America. Thus, we added a shared screen that was placed between the human and the robot. On that screen, the world map, as well as the remaining time and the current score, was displayed. As soon as the Matcher made a selection, it was highlighted on the shared map in bright green. 

One challenge with visualizing the Matcher's selection was that in some pilot tests we saw that it would lead pairs to just pick any country in the rough region and incrementally take it from there. This made the game too easy and less fun to play. Thus, we restricted the Matcher to only make two guesses per target country. If the first selection was correct, the team would score 2 points. If the initial selection was incorrect, the Director could continue describing and, if the updated Matcher selection was correct, they would still score 1 point. Only if the second selection was still wrong, they wouldn't score at all. While this slightly changes the dynamic of the game and adds an element of incremental correction, the Director is still required to give a complete description of the country before the Matcher would reasonably make a guess about the target country.

\subsection{Agent Implementation}
\label{subsec:neil:agentimplementation}
We used the blended robot platform Furhat \cite{c5} as an embodiment for our agent (cf~Figure \ref{fig:robot_setup}). Furhat is equipped with a rigid mask of a male face on which a facial texture is projected from within. The robot has two motors to control the head's tilt and pan. Due to the male appearance of the Furhat mask, we used the \newcite{Cereproc} voice William which has a similar level of expressiveness as the Kate voice. The interaction was again remote-controlled by a Wizard who was placed behind a curtain. The Wizard interface was similar to the one used in the web-based experiment (cf~Figure~\ref{fig:wizardinterface} right). However, due to the experience gained from the online study, we changed the wording of some of the utterances to be less ambiguous and added 3 buttons to solve common game situations that the agent could not react to appropriately. In addition, we added two happy and two sad vocal reactions the Wizard could use in case a selection was correct or false. To allow for repeated game plays, the robot could also mention that it did or did not remember countries mentioned by the Director.

The embodied version of the agent allows to incorporate head and gaze movements as well as facial expressions. Facial expressions were mostly used when the robot gave a happy or sad reaction to a correct or false guess. However, the Wizard could also control them independently to use in other game situations. The gaze of the robot was mostly focused on the shared screen. By directing the head orientation and gaze towards the left of right, the robot could indicate that it is looking at the Americas or Asia and Oceania. Occasionally and when the Director had longer thinking pauses, the Wizard could briefly direct the robot's gaze towards the user and then back to the game.
To account for the robot being based in Sweden and the robot's past travel history (which was mentioned in one of the post-game interactions), Sweden and France were added to the list of countries known to the agent from the start of the game.

\subsection{Data Collection}
\label{subsec:neil:datacollection}

\begin{figure}[t]
    \begin{center} 
    \includegraphics[width=\linewidth]{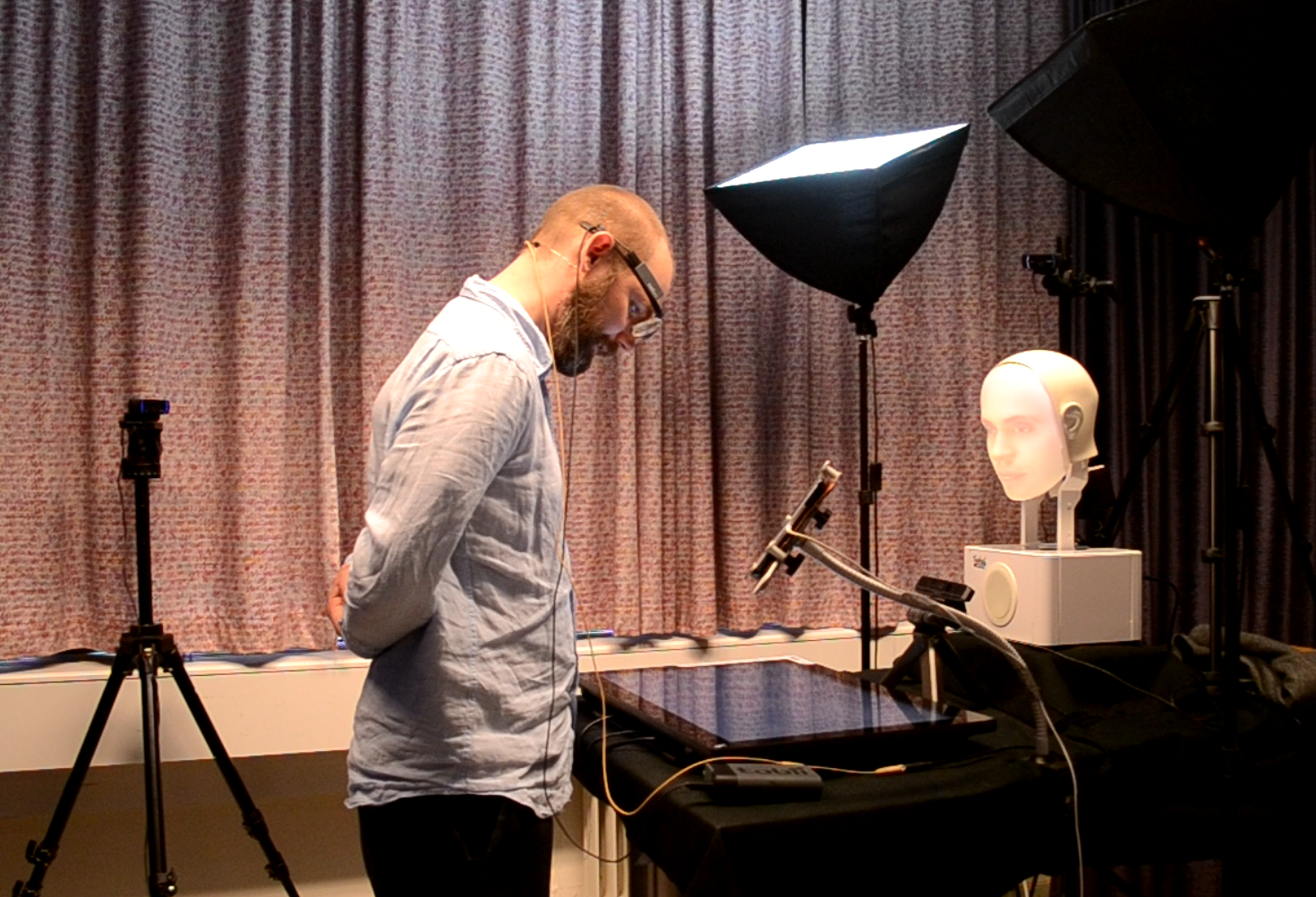}
    
    \caption{The experimental setup for the data collection with the embodied Furhat robot Neil.}
    \label{fig:robot_setup}
    \end{center}
\end{figure}

We recruited 60 participants at Uppsala University to take part in the data collection. Participants received course credits for their participation. 2 participants were excluded because they suspected the agent to be remote controlled or had technical failures. Out of the remaining 58 participants, 49 had three game interactions with the robot, 7 had two interactions, and 2 only played once with the agent. Participants had at least 24 hours of break between the robot interactions. On average, participants in the embodied data collection were younger than in the web-based study (\me = 24.5, \sd = 4.5) and all except of two had a background in Computer Science. 18 participants identified as female, one chose not to disclose this information and the remaining identified as male.

The setup of the experiment space is as shown in Figure \ref{fig:robot_setup}. 
At the beginning of their first session with the robot, participants gave informed consent, read the game rules, and filled out a short demographic survey. After each game interaction, they were given a survey similar to the one used in the online data collection to measure people's perception of the agent and their satisfaction with the game interaction. Since this data collection with the robot was part of a larger study, participants had an additional short interaction before and after they played the game with the robot. As part of this conversation, the robot asked participants whether they believed they learned something during the game and whether they believe they will remember the countries next time they play. In the conversation participants had with the robot prior to playing the first game, the robot asked how they would rate their own geography skills. 30 replied that their knowledge is rather good, while 28 subjectively rated their knowledge as poor.


\section{Results}
\label{sec:results}

The corpora we collected consist of 48 game interactions and 8 hours of recorded audio in the web-based corpus and 161 game interactions and 26.8 hours of recorded audio in the embodied-interaction corpus. 
In this paper, we focus on discussing the results of the perceived engagement in the game and the subjective rating of the learning gain. 
Since the gameplay and agent utterances were changed between the web-based and embodied interaction, we cannot directly compare the results between the two. Hence, we report them separately.

\subsection{Web-Based Data Collection}

All participants were able to score points when playing with the agent. The score ranged between 3 and 29, with an average of 14.44 points (\sd = 6.44). In other words, on average participants were able to score more than one point per minute.

In the post-game questionnaire, we asked participants a variety of questions regarding their interaction with the robot and their experience with the game on a five-point Likert scale. People generally liked working with the agent (\me = 3.35, \sd = 1.26) and found it relatively easy to play the game with it (\me = 3.38, \sd = 1.33). They also found it easy to understand the agent (\me = 4.13, \sd = 1.04) and reported they could talk to the agent similar to how they would talk to another human (\me = 3.92, \sd = 1.03). While this could be an indication that people have figured out the agent was controlled by a human, many reported in the free-text fields that they were talking to an agent. Thus, it seems that people found the interaction to be quite natural despite their believe they were talking to an autonomous agent. 

\begin{figure}[t]
    \begin{center} 
    \includegraphics[width=\linewidth]{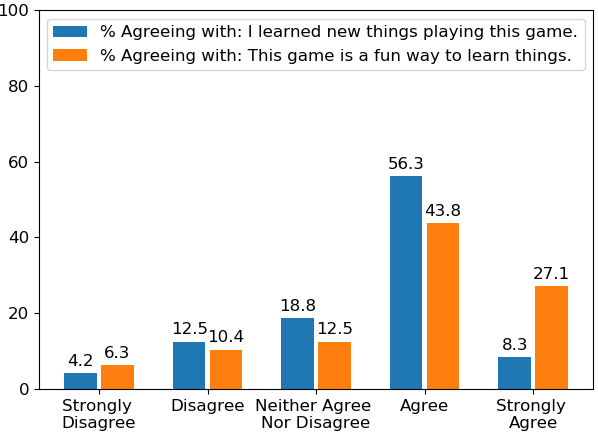}
    
    \caption{Participant's subjective evaluation of the learning value of the web-based version of the game.}
    \label{fig:results_online}
    \end{center}
\end{figure}

The aim of the RDG-Map game is not just to engage people in an interaction with an artificial agent, we also hope to create an environment in which people learn more about the world map. As shown in Figure \ref{fig:results_online}, 64.6\% of the participants reported they have learned something playing the game (\me = 3.52, \sd = 0.97) and 70.8\% of participants found the game to be a fun way to learn new things (\me = 3.75, \sd = 1.16). Even though these ratings are clearly subjective, it gives a first indication that the game could increase the participant's geographic literacy. 
Participants who were critical regarding the learning aspect of the game were mostly unsatisfied with the robotic and/or repetitive interaction with the agent. One participant said: ``The partner was robotic. It was not fun''. Another, who reported she hadn't learned anything but found the game in general a fun way of learning something wrote: ``I think it was it was jusr (sic) frustrating when the other party does not know where things are.'' Some participants wished the interaction would be less repetitive: ``After a while, you start to repeat the same formula of a description to the player, so it gets repetitive.'' Another suggested: ``I feel that the other side should have more responses available in order to better describe locations effectively.''

Most participants, however, were positive about the learning aspect of the game. One participant wrote: ``I think its (sic) a very interactive learning tool which makes studying more enjoyable.'' Another pointed out that ``it allows you to learn the world map in a fun way by learning to describe it.'' It is noteworthy that even the older generation had a positive experience with the game. A 55-year-old female participant said: ``it was fun as I got to learn where some countries I had never heard of before were''. One participant with a (self-reported) Master's degree even wrote: ``i learned more than i have in years.''

\subsection{Embodied Data Collection}

Similar to the online data collection, participants were able to understand the robot very well over all three of the interactions they had with the robot (1st game: \me = 4.58, \sd = 0.5, 2nd game: \me = 4.57, \sd = 0.5, 3rd game: \me = 4.3, \sd = 0.58). They also mainly agreed that they talked to the robot in the way they would talk to another human (1st game: \me = 3.77, \sd = 1.05, 2nd game: \me = 3.8, \sd = 0.98, 3rd game: \me = 3.55, \sd = 1.02). 

\begin{figure}[t]
    \begin{center} 
    \includegraphics[width=\linewidth]{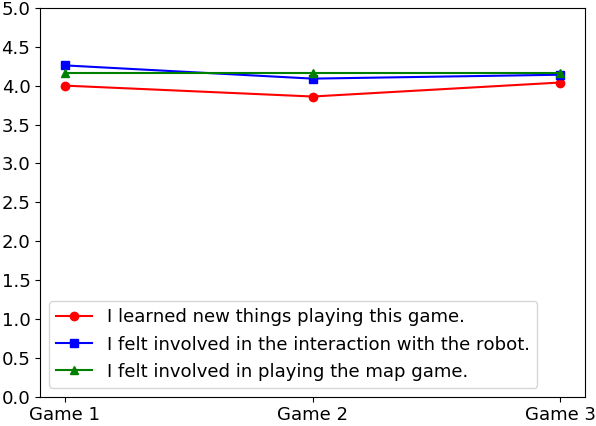}
    
    \caption{Participant's subjective evaluation of the learning value and their engagement when playing the game with the embodied agent.}
    \label{fig:results_embodied}
    \end{center}
\end{figure}

Participants felt very engaged both with the robot (1st game: \me = 4.26, \sd = 0.55, 2nd game: \me = 4.09, \sd = 0.64, 3rd game: \me = 4.14, \sd = 0.65) and the game itself (1st game: \me = 4.16, \sd = 0.75, 2nd game: \me = 4.16, \sd = 0.71, 3rd game: \me = 4.16, \sd = 0.62) and this engagement did not decrease with having repeated interactions (cf~Figure~\ref{fig:results_embodied}). ANOVA analysis revealed that repeated interactions had no significant influence on the perceived engagement with the robot, $F(1,160) = 1.121, p = .291$, and the game, $F(1,160) = 0.002, p = .968$.  This is important since it suggests that the engagement does not come from the pure novelty of the game but rather from the game and interaction dynamics. When asked by the robot after the first game, all 58 participants reported they enjoyed playing the game.

We are also interested in understanding whether participants believe the game increases their geographic literacy. Participants overall agreed to the statement that they learned new things playing this game (1st game: \me = 4.0, \sd = 0.73, 2nd game: \me = 3.86, \sd = 0.77, 3rd game: \me = 4.04, \sd = 0.73). It is noteworthy that the agreement with this statement did not decrease after the repeated interactions, $F(1,160) = 0.049, p = .825$, suggesting that this game can increase participant's knowledge even when playing it multiple times (cf~Figure~\ref{fig:results_embodied}). 
When asked by the robot, all except for one participant reported they believed they learned something when playing the game. At the beginning of the second session, the robot asked participants whether they remember some of the countries from the previous game. Out of all 55 participants who had this conversation, only four reported they don't recall anything. 

On inquiry, 38 of 51 participants told the robot they found it easier to play the second game round compared to the first time they played. This is in line with the game scores: While participants on average scored 22.24 points (\sd = 9.58) in the first game, they scored 26.96 (\sd = 10.6) in the second and 29.0 (\sd = 10.68) in the third game. The higher game score in the embodied compared to the web-based interaction likely comes from the different scoring system that allowed participants to score up to two points per target.


\section{Discussion \& Future work}
\label{sec:discussion}

In this paper, we introduced the RDG-Map game, which was developed to offer engaging interactions with a dialogue agent while increasing participant's geographic literacy. 
In our two data collections, we showed that participants had fun playing the game both when playing with an unembodied agent on the web and with a robot in a classroom setting. 
Most players also reported they learned something by playing the game. 
According to the participants in the embodied data collection, they were even able to recall some of the information after a couple of days when they came back to play the game again. \textit{Thus, our results are a first indication that the game could increase the participant's geographic literacy}. 
One shortcoming of the data collection presented in this work is that we did not measure the objective learning gain in the game. 
We experimented with having a pre- and post-game test of geographic knowledge in the web-based version of the game. 
However, we could not make sure that our test data are reliable since it would be very easy for participants to look up the correct answer to the questions on the internet. 
Since the game in the embodied interaction was part of a larger study setup, it was not possible to add a pre- and post-game test to the experiment. 
In the future, we will pursue another experiment in which we explicitly measure the learning gain of our game in comparison to other ways of increasing geographic literacy. 

We found many participants playing with the embodied agent to be anxious when learning that they would play a game about the world map. They seemed to be embarrassed about their lack of knowledge about countries in the world. Giving participants the role of the Director and providing all the required information on the world map turned out to be reassuring for them and lower their reluctance to play the game. 
\textit{By taking the role of the tutor in this interaction, they could gain knowledge in the absence of any judgment of their own skill level}. 
When playing with the agent, many were surprised how little knowledge it had about the world map. While some find this annoying, it can be reassuring to others because their own lack of knowledge seems less humiliating. 
In the future, we plan to implement the game so the agent can play both the role of the Matcher and the Director. 
With the experience we gained in this data collection, we believe that always assigning the role of the Director to the human first could raise participant's self-confidence when playing with the agent. \textit{The results presented in this paper also show that the game keeps being engaging and increase the subjective geography knowledge even in repeated interactions}. By providing a web-based version of the game, we can allow people to play the game and learn the location of countries in the world at their own pace and as many times as possible without the fear of judgment when they don't score as high as others. 

We are currently in the process of transcribing and annotating all interactions with domain-specific speech acts. We intend to release this corpus in the near future. 
In addition, we use the corpus to train an agent that can play the game in both the role of the Matcher and the Director fully autonomously.
When it comes to implementing the agent, the focus will not merely be on creating the perfect SDS, which scores as many points as possible in the game. Instead, the agent should try to both optimize the engagement and the learning outcome of the game. Such an agent could, for example, repeat the name of the country multiple times in order to confirm recently acquired knowledge. This is particularly challenging for the dialogue systems as they're typically designed to maximize task performance \cite{paetzel2015so}. As shown in Figure~\ref{fig:map-dir}, descriptions of countries can become quite complex. Often people first establish other anchoring points that are close to the target country and then describe the target starting from the newly established anchor. Conventional language understanding approaches will not be able to resolve such complex descriptions. 
\textit{With the RDG-Map domain and our efforts to create an SDS that can play the game autonomously, we are ultimately trying to add real and natural interactivity to the genre of serious games}. We hope that the knowledge gained in such games can have a lasting impact on people's geographic literacy, a crucial skill to participate and make informed (political) decisions in today's globalized world.


\section{Conclusion}
\label{sec:conclusion}
In this work, we presented the RDG-Map game, which was developed to engage people in learning about the world map. We implemented two versions of the game, one web-based and one involving an embodied agent. For both versions, we discussed initial results from two large data collections involving a Wizard controlled agent playing with human game partners. These results suggest that people felt engaged both with the agent and the game itself. In addition, participants reported they learned about locations of countries in the world playing the game, and they recall some of the information even days after the interaction.

\bibliography{acl2018}

\begin{thebibliography}{31}
\expandafter\ifx\csname natexlab\endcsname\relax\def\natexlab#1{#1}\fi

\bibitem[{Al~Moubayed et~al.(2012)Al~Moubayed, Beskow, Skantze, and
  Granstr{\"o}m}]{c5}
Samer Al~Moubayed, Jonas Beskow, Gabriel Skantze, and Bj{\"o}rn Granstr{\"o}m.
  2012.
\newblock {Furhat: a back-projected human-like robot head for multiparty
  human-machine interaction}.
\newblock In \emph{Cognitive Behavioural Systems}, pages 114--130. Springer.

\bibitem[{Anderson et~al.(2019)Anderson, Perrin, Jiang, and
  Madhumitha}]{anderson201913}
Monica Anderson, Andrew Perrin, Jingjing Jiang, and Kumar Madhumitha. 2019.
\newblock \href
  {https://www.pewresearch.org/fact-tank/2019/04/22/some-americans-dont-use-the-internet-who-are-they}
  {10\% of americans don’t use the internet. who are they}.
\newblock \emph{Pew Research Center}.

\bibitem[{Baker(2016)}]{baker2016stupid}
Ryan~S Baker. 2016.
\newblock Stupid tutoring systems, intelligent humans.
\newblock \emph{International Journal of Artificial Intelligence in Education},
  26(2):600--614.

\bibitem[{Cereproc()}]{Cereproc}
Cereproc.
\newblock https://www.cereproc.com/.
\newblock https://www.cereproc.com.
\newblock 2019.

\bibitem[{CFK and Geographic(2006)}]{natgeo:2006}
Roper CFK and National Geographic. 2006.
\newblock \href
  {https://media.nationalgeographic.org/assets/file/NGS-Roper-2006-Report.pdf?fbclid=IwAR0x1Y1UEkr0MVvZHOfIfs3vmERIc0XMG1rHsBW0JAnbHFECUAIbJ1AswNk}
  {{Geographic Literacy Survey}}.

\bibitem[{CFR and Geographic(2016)}]{natgeo:2016}
CFR and National Geographic. 2016.
\newblock {What College-Aged Students Know About the World, A survey on Global
  Literacy}.

\bibitem[{Craig et~al.(2013)Craig, Hu, Graesser, Bargagliotti, Sterbinsky,
  Cheney, and Okwumabua}]{craig2013impact}
Scotty~D Craig, Xiangen Hu, Arthur~C Graesser, Anna~E Bargagliotti, Allan
  Sterbinsky, Kyle~R Cheney, and Theresa Okwumabua. 2013.
\newblock The impact of a technology-based mathematics after-school program
  using aleks on student's knowledge and behaviors.
\newblock \emph{Computers \& Education}, 68:495--504.

\bibitem[{De~Vries et~al.(2017)De~Vries, Strub, Chandar, Pietquin, Larochelle,
  and Courville}]{de2017guesswhat}
Harm De~Vries, Florian Strub, Sarath Chandar, Olivier Pietquin, Hugo
  Larochelle, and Aaron Courville. 2017.
\newblock Guesswhat?! visual object discovery through multi-modal dialogue.
\newblock In \emph{Proceedings of the IEEE Conference on Computer Vision and
  Pattern Recognition}, pages 5503--5512.

\bibitem[{Freedman(1999)}]{freedman1999atlas}
Reva Freedman. 1999.
\newblock Atlas: A plan manager for mixed-initiative, multimodal dialogue.
\newblock In \emph{AAAI-99 workshop on mixed-initiative intelligence}, pages
  1--8.

\bibitem[{Gertner and VanLehn(2000)}]{gertner2000andes}
Abigail~S Gertner and Kurt VanLehn. 2000.
\newblock Andes: A coached problem solving environment for physics.
\newblock In \emph{International conference on intelligent tutoring systems},
  pages 133--142. Springer.

\bibitem[{Graesser(2016)}]{graesser2016conversations}
Arthur~C Graesser. 2016.
\newblock Conversations with autotutor help students learn.
\newblock \emph{International Journal of Artificial Intelligence in Education},
  26(1):124--132.

\bibitem[{Graesser et~al.(2004)Graesser, Lu, Jackson, Mitchell, Ventura, Olney,
  and Louwerse}]{graesser2004autotutor}
Arthur~C Graesser, Shulan Lu, George~Tanner Jackson, Heather~Hite Mitchell,
  Mathew Ventura, Andrew Olney, and Max~M Louwerse. 2004.
\newblock Autotutor: A tutor with dialogue in natural language.
\newblock \emph{Behavior Research Methods, Instruments, \& Computers},
  36(2):180--192.

\bibitem[{Graesser et~al.(2001)Graesser, VanLehn, Ros{\'e}, Jordan, and
  Harter}]{graesser2001intelligent}
Arthur~C Graesser, Kurt VanLehn, Carolyn~P Ros{\'e}, Pamela~W Jordan, and Derek
  Harter. 2001.
\newblock Intelligent tutoring systems with conversational dialogue.
\newblock \emph{AI magazine}, 22(4):39--39.

\bibitem[{Johnson et~al.(2005)Johnson, Vilhj{\'a}lmsson, and
  Marsella}]{johnson2005serious}
W~Lewis Johnson, Hannes~H{\"o}gni Vilhj{\'a}lmsson, and Stacy Marsella. 2005.
\newblock Serious games for language learning: How much game, how much ai?
\newblock In \emph{AIED}, volume 125, pages 306--313.

\bibitem[{Koedinger et~al.(1997)Koedinger, Anderson, Hadley, and
  Mark}]{koedinger1997intelligent}
Kenneth~R Koedinger, John~R Anderson, William~H Hadley, and Mary~A Mark. 1997.
\newblock Intelligent tutoring goes to school in the big city.

\bibitem[{Koedinger et~al.(2013)Koedinger, Stamper, McLaughlin, and
  Nixon}]{koedinger2013using}
Kenneth~R Koedinger, John~C Stamper, Elizabeth~A McLaughlin, and Tristan Nixon.
  2013.
\newblock Using data-driven discovery of better student models to improve
  student learning.
\newblock In \emph{International Conference on Artificial Intelligence in
  Education}, pages 421--430. Springer.

\bibitem[{Lesgold et~al.(1988)}]{lesgold1988sherlock}
Alan Lesgold et~al. 1988.
\newblock Sherlock: A coached practice environment for an electronics
  troubleshooting job.

\bibitem[{Litman and Silliman(2004)}]{litman2004itspoke}
Diane~J Litman and Scott Silliman. 2004.
\newblock Itspoke: An intelligent tutoring spoken dialogue system.
\newblock In \emph{Demonstration papers at HLT-NAACL 2004}, pages 5--8.
  Association for Computational Linguistics.

\bibitem[{Manuvinakurike and DeVault(2015)}]{manuvinakurike2015pair}
Ramesh Manuvinakurike and David DeVault. 2015.
\newblock Pair me up: A web framework for crowd-sourced spoken dialogue
  collection.
\newblock In \emph{Natural Language Dialog Systems and Intelligent Assistants},
  pages 189--201. Springer.

\bibitem[{Manuvinakurike et~al.(2016)Manuvinakurike, Kennington, DeVault, and
  Schlangen}]{manuvinakurike2016real}
Ramesh Manuvinakurike, Casey Kennington, David DeVault, and David Schlangen.
  2016.
\newblock Real-time understanding of complex discriminative scene descriptions.
\newblock In \emph{Proceedings of the 17th Annual Meeting of the Special
  Interest Group on Discourse and Dialogue}, pages 232--241.

\bibitem[{Manuvinakurike et~al.(2015)Manuvinakurike, Paetzel, and
  DeVault}]{manuvinakurike2015reducing}
Ramesh Manuvinakurike, Maike Paetzel, and David DeVault. 2015.
\newblock Reducing the cost of dialogue system training and evaluation with
  online, crowd-sourced dialogue data collection.
\newblock \emph{SEMDIAL 2015 goDIAL}, page 113.

\bibitem[{McNamara et~al.(2010)McNamara, Jackson, and
  Graesser}]{mcnamara2010intelligent}
Danielle~S McNamara, G~Tanner Jackson, and Art Graesser. 2010.
\newblock Intelligent tutoring and games (itag).
\newblock In \emph{Gaming for classroom-based learning: Digital role playing as
  a motivator of study}, pages 44--65. IGI Global.

\bibitem[{McNamara et~al.(2004)McNamara, Levinstein, and
  Boonthum}]{mcnamara2004istart}
Danielle~S McNamara, Irwin~B Levinstein, and Chutima Boonthum. 2004.
\newblock istart: Interactive strategy training for active reading and
  thinking.
\newblock \emph{Behavior Research Methods, Instruments, \& Computers},
  36(2):222--233.

\bibitem[{Michael and Chen(2005)}]{michael2005serious}
David~R Michael and Sandra~L Chen. 2005.
\newblock \emph{Serious games: Games that educate, train, and inform}.
\newblock Muska \& Lipman/Premier-Trade.

\bibitem[{Mitros and Sun(2014)}]{mitros2014creating}
Piotr Mitros and Felix Sun. 2014.
\newblock Creating educational resources at scale.
\newblock In \emph{2014 IEEE 14th International Conference on Advanced Learning
  Technologies}, pages 16--18. IEEE.

\bibitem[{Mitrovic and Ohlsson(1999)}]{mitrovic1999evaluation}
Antonija Mitrovic and Stellan Ohlsson. 1999.
\newblock Evaluation of a constraint-based tutor for a database language.

\bibitem[{Paetzel et~al.(2015)Paetzel, Manuvinakurike, and
  DeVault}]{paetzel2015so}
Maike Paetzel, Ramesh Manuvinakurike, and David DeVault. 2015.
\newblock {``So, which one is it?'' The effect of alternative incremental
  architectures in a highperformance game-playing agent}.
\newblock In \emph{The 16th Annual Meeting of the Special Interest Group on
  Discourse and Dialogue (SigDial)}.

\bibitem[{Paetzel et~al.(2014)Paetzel, Racca, and
  DeVault}]{paetzel2014multimodal}
Maike Paetzel, David~Nicolas Racca, and David DeVault. 2014.
\newblock A multimodal corpus of rapid dialogue games.
\newblock In \emph{LREC}, pages 4189--4195.

\bibitem[{Pane et~al.(2014)Pane, Griffin, McCaffrey, and
  Karam}]{pane2014effectiveness}
John~F Pane, Beth~Ann Griffin, Daniel~F McCaffrey, and Rita Karam. 2014.
\newblock Effectiveness of cognitive tutor algebra i at scale.
\newblock \emph{Educational Evaluation and Policy Analysis}, 36(2):127--144.

\bibitem[{Trinh et~al.(2017)Trinh, Asadi, Edge, and
  Bickmore}]{trinh2017robocop}
Ha~Trinh, Reza Asadi, Darren Edge, and T~Bickmore. 2017.
\newblock Robocop: A robotic coach for oral presentations.
\newblock \emph{Proceedings of the ACM on Interactive, Mobile, Wearable and
  Ubiquitous Technologies}, 1(2):27.

\bibitem[{Zarrie{\ss} et~al.(2016)Zarrie{\ss}, Hough, Kennington,
  Manuvinakurike, DeVault, Fern{\'a}ndez, and Schlangen}]{zarriess2016pentoref}
Sina Zarrie{\ss}, Julian Hough, Casey Kennington, Ramesh Manuvinakurike, David
  DeVault, Raquel Fern{\'a}ndez, and David Schlangen. 2016.
\newblock Pentoref: A corpus of spoken references in task-oriented dialogues.
\newblock In \emph{10th edition of the Language Resources and Evaluation
  Conference}.

\end{thebibliography}
\bibliographystyle{acl_natbib}

\end{document}